\documentclass{article}
\usepackage{arxiv}

\usepackage{amsmath}
\usepackage{url}
\usepackage[pdftex]{graphicx}
\usepackage{cite}
\usepackage[utf8]{inputenc}
\usepackage[T1]{fontenc}
\usepackage[english]{babel}
\usepackage{color}

\usepackage{lmodern} 
\usepackage[T1]{fontenc}
\begin{document}

\title{A Single Scalable LSTM Model for Short-Term Forecasting of Massive Electricity Time-Series}

\author{Andr\'es~M.~Alonso, Francisco J.~Nogales, 
        and~Carlos~Ruiz
\thanks{A. M. Alonso, F. J. Nogales and C. Ruiz are with the Department
of Statistics and the UC3M-Santander Big Data Institute (IBiDat), Universidad Carlos III de Madrid, Avda. de
la Universidad 30, 28911-Leganés, Spain. (e-mails: andres.alonso@uc3m.es; fcojavier.nogales@uc3m.es; carlos.ruiz@uc3m.es)}
}


\maketitle

\begin{abstract}
Most electricity systems worldwide are deploying advanced metering infrastructures to collect relevant operational data. In particular, smart meters allow tracking electricity load consumption at a very disaggregated level and at high frequency rates. This data opens the possibility of developing new forecasting models with a potential positive impact in electricity systems. We present a general methodology that is able to process and forecast a large number of smart meter time series. Instead of using traditional and univariate approaches, we develop a single but complex recurrent neural-network model with long short-term memory that can capture individual consumption patterns and also consumptions from different households. The resulting model can accurately predict future loads (short-term) of individual consumers, even if these were not included in the original training set. This entails a great potential for large scale applications as once the single network is trained, accurate individual forecast for new consumers can be obtained at almost no computational cost. The proposed model is tested under a large set of numerical experiments by using a real-world dataset with thousands of disaggregated electricity consumption time series. Furthermore, we explore how geo-demographic segmentation of consumers may impact the forecasting accuracy of the model.

\end{abstract}


\section{Introduction}
Most electricity systems worldwide are undergoing major transformations driven by the need of improving their sustainability and efficiency. These affect both their technical and economical operation and pose new challenges for all the different agents that participate in the electricity supply chain: generation, transportation, distribution and consumption. 
One key aspect of these transformations is the deployment of advanced metering infrastructure (AMI) technologies, which are being integrated in most electricity systems worldwide  \cite{depuru2011smart}. In particular smart meters are being progressively installed in individual households and allow tracking electricity consumption dynamics at a very disaggregated level and high frequency rates. 

Moreover, since the recent deregulation of many electricity markets worldwide, there are several agents (market operators, gencos, retailers, consumers, etc.) that are highly interested on obtaining the maximum value from this type of data. In this sense, new business opportunities arise that are linked to the application and development of state-of-the-art analytics and data science techniques. \cite{yildiz2017recent}.

Specifically, developing an accurate and computationally efficient short-term forecasting model for individual consumers (smart meters) has several important applications within the new smart grid paradigm. One of them relates to the successful implementation of demand response (DR) policies. Once real time price tariffs are fully deployed in distribution systems, the forecasting model can learn from historical consumption patterns, as well as from each individual consumers’ reaction to prices (elasticity). This model can then be used to anticipate (forecast) the impact of different pricing strategies in both aggregated and disaggregated loads, and design these to better benefit the system. In this sense it can be helpful to identify those consumers more suitable for DR policies, anticipate and prevent systems peaks or congested lines, adapt load consumption to renewable generation availability, identify consumption anomalies that may originate from equipment failure or from electricity theft, etc. 

Furthermore, smart-meter datasets may help to better understand which are the main factors that drive electricity consumption  \cite{wang2018review}, and how these can be used to improve the system's efficiency. For instance, electricity retailers might be interested in precise load forecasts for optimal participation in futures and spot markets \cite{conejo2010decision}. Similarly, distribution companies might benefit from this sort of predictions to improve the network reliability.

For these reasons, the time series recorded by smart meters open the possibility of improving the existing forecasting models for electricity loads, from an entire system to a single household. 

Until recently, most of the models proposed in the literature, and used by the industry, have focused on forecasting aggregated loads at a system or substation level. However, these models are not appropriate for dealing with time series from more disaggregated loads (buildings or individual households), \cite{hong2016probabilistic}. For instance, at a household level, time series are very volatile, include a high amount of noise, and are unevenly (and nonlinearly) affected by calendar effects and meteorological variables. 
This has motivated the used of new modeling techniques, especially in the field of Machine Learning (ML), that are able to capture potential nonlinearities among the variables, do not assume any data generation process (non-parametric), and can exploit these large datasets for their training. 
In particular, Artificial Neural Networks (ANN) based models have proven effective in this type of context. 
One of their potential advantages, compared to traditional time series approaches, is to be able to jointly treat and combine information from several time series to improve forecasting accuracy, i.e., without studying the series in isolation \cite{bandara2017forecasting}.

In this work we will focus on a particular ANN network architecture: Recurrent Neural Networks (RNN) with Long Short-term Memory (LSTM) \cite{hochreiter1997long}. These are specifically designed to capture sequential dependencies between the data, as is the case of time series, and information from different consumers' dynamics. 
However, the main drawback of these models is that, in order to be effective, they demand a very high computational burden for their training. This may hinder their practical implementation in a smart meter framework, as this application involves processing and forecasting up to hundreds of thousands of time series. 

To overcome this difficulty, in this work we propose a new forecasting approach to deal effectively with a large number of time series, as is the case of smart meter data. In particular, we propose to train a single RNN-LSTM model over a subset of the available smart meter time series. We show how, after an appropriate training and parameter tuning, the resulting model can accurately predict future loads of individual consumers, even if these were not included in the original training set. Hence, the resulting tool has a great potential for large scale applications (Big Data) as once the single network is trained, accurate individual forecast can be obtained at almost no computational cost. We test the validity of our approach under an extensive set of numerical experiments based on a real-world dataset that includes several thousand of household load time series. Results indicate that the proposed methodology is able to improve the forecasting accuracy of relevant benchmarks and also benefit from geo-demographic segmentation of consumers in the dataset.

\subsection{Literature Review}

There are several works that have focused on comparing the performance of different forecasting techniques in the context of smart meter electricity consumption data. For instance, 
reference \cite{edwards2012predicting} compares alternative ML techniques to predict the hourly consumption loads of three residential households. A large battery of simulations suggest that the best forecasting technique depends on the particular household under study, although Least Squares Support Vector Machines shows overall good performance. 
A review of forecasting techniques, based on Artificial Intelligence, for electrical load demands is presented in \cite{raza2015review}. It shows that ANNs have a great potential, but present challenges related to initialization parameters, slow convergence or local minima. In this vein, the authors in \cite{ma2017modeling} revise several load forecasting techniques in the context of distributed energy systems. They conclude that the existing methodologies present either low forecast accuracy or high computational burden.

With the same aim, \cite{gajowniczek2017electricity} tests the performance of several techniques (based both on classical time series and ML) to forecast individual household consumptions. Apart from considering standard predictors based on historical loads, meteorological variables and calendar effects, the authors incorporate household activity patterns, which enhance the accuracy of the predictions. 
Similar conclusions are obtained  in \cite{hsiao2014household}, where the behavior patterns are also considered to improve the predictive accuracy of individual household electricity demands. 
Moreover, \cite{sevlian2018scaling} characterizes empirically how the forecasting accuracy of different techniques varies with the level of aggregation of the series.

Other works seek to procure probabilistic forecasts for smart meter data, in this case \cite{taieb2016forecasting} presents an additive quantile regression model that incorporates a boosting procedure for variable selection. The model outperforms standard probabilistic approaches for aggregated data. In a recent work and in a similar application, \cite{taieb2019hierarchical} proposes a methodology to compute coherent probabilistic forecasts for hierarchical time series. 
A predictive methodology based on LASSO is proposed in \cite{li2017sparse} to deal with sparsity and select relevant lag orders for autoregressive models. The model is applied to individual consumption data and its performance is further enhanced by including consumption series from other users. 

With the recent advent of cloud and distributed computing, computationally intensive techniques such as Deep Learning (DL), and in particular RNN, have been applied to forecast disaggregated electricity loads.

With the idea of exploiting similarities between time series, \cite{bandara2017forecasting} proposes a forecasting framework that combines clustering and recurrent neural networks. In particular, first subgroups of time series are created based on cross-similarities and then a predictive LSTM network is trained on each subgroup. The model is tested under time series from the banking industry (monthly) a withdrawals from  automatic teller machines (daily). 
A similar idea in the context of load forecasting is employed by \cite{quilumba2014using}, where clusters of consumers with similar load patterns are formed before the adjusting a NN forecasting model for the aggregated series. 
A Self-Recurrent Wavelet Neural Network is used in \cite{chitsaz2015short} to forecast the load of buildings in microgrids. The predictive tool makes used of feedback loops to better deal with volatile time series.
%
Moreover,  \cite{tascikaraoglu2016short} shows how an approach incorporating temporal and spatial  information, can be used to identify relevant features among different households and improve the forecasting accuracy. Reference \cite{yildiz2018household} also employs a clustering analysis to extract typical daily consumption patterns that can be used to improve the accuracy of the forecasting tool.

At a district level, \cite{ahmad2019deep} addresses the problem of accurate short-term load forecasting including both meteorological and technical variables. The proposed model is based on a DL algorithm that combines different back-propagation techniques to ease its computational burden. 
Similarly, a DL forecasting model for individual consumption time series, based on Conditional Restricted Boltzmann Machine, is presented in \cite{mocanu2016deep}. The results, tested on a single consumer historical data set, show that it outperforms ANN, Support Vector Machines and RNN techniques. 
With the same aim, \cite{shi2017deep} presents a pooling-based deep recurrent
 neural network designed to avoid over-fitting. The method outperforms traditional ARIMA, Support Vector Regression and RNN. 

 A deep neural network model based on LSTM architecture is used in \cite{kong2017short} to forecast residential loads. It is shown how, in order to forecast aggregated loads, in may be convenient to aggregate individual forecasts. 
Finally, another LSTM model is adapted in  \cite{wang2019probabilistic}  to provide probabilistic forecasts for household loads. To this end, a pinball loss function is used to train the model by evaluating the quantile errors. 

\subsection{Contributions}
To the best of the authors' knowledge, this is the first work that aims to develop an efficient forecasting tool for high frequency and disaggregated loads, making an emphasis on its scalability, and hence potential application for massive (hundreds of thousand) smart-meter time-series. It is worth noting that the proposed pooling based method and RNN model share some common features with the ones proposed in \cite{shi2017deep} and \cite{bandara2017forecasting}, respectively. However, these are extended to work with high frequency time series, to account for exogenous variables, that enrich the forecasting model, and to provide accurate forecast for out-of-sample consumers.  

To summarize, considering the state of the art described in the previous section, 
the main contributions of this work are three-fold:
\begin{enumerate}
	\item To make use of a single LSTM model to produce accurate forecasts for general high-frequency time series, in our case households' load smart-meters.
	\item To build the LSTM model so that it can be trained efficiently using only a subset of the smart meters, making it scalable and suitable to deal with massive time-series (Big Data).
	\item To validate the proposed methodology with a real-world dataset involving thousands of smart meters and by analyzing how geo-demographic segmentation can impact forecasting accuracy, and improving the out-of-sample performance over relevant benchmarks.
\end{enumerate}

With these contributions, the proposed forecasting model can be efficiently trained to provide accurate near real-time forecast for hundreds of thousands of smart meters. As we show in this work, training a unique LSTM network, which is the most demanding computational task, would need to be done only once every few months, and by using just a representative subset of the smart meters time series. Once the network is trained, and its input properly defined, it can provide short-term smart meters’ forecast at almost no computational costs, as the next sections show.

\subsection{Paper Organization}
The rest of the manuscript is organized as follows. In Section \ref{DA}, we describe the real-world dataset and summarize the main properties of the associated time series. In Section \ref{Met}, the proposed methodology based on RNN is introduced together with the numerical results obtained in an extensive back-testing including relevant benchmarks. Finally, Section \ref{Con} concludes by emphasizing the potential of the proposed methodology in large-scale applications. 

\section{Data and Descriptive Analysis}\label{DA} 

We have accessed the public energy consumption registers from \cite{dataLondon}. In particular, this dataset contains a sample of 5,567 London households with the corresponding energy consumption, in kWh (per half hour), for each unique household identifier, date and time, and CACI ACORN categories (classification of residential neighborhoods). There are six categories that provide a geo-demographic segmentation of London's population. Each of these six categories are further divided into a total of 18 groups, to increase homogeneity, and our methodology will be applied to each of these groups. More details about CACI ACORN classification can be found in \cite{CACI}.

The proposed methodology is general and deals with high-frequency time series. Our dataset contains a frequency of half hour, but we have aggregated it into an hourly frequency to reduce a bit the dimension and the associated volatility. That means each smart meter in each of the 18 groups contains 8760 hours corresponding to its energy consumption, in kWh. Note that, in any case, our methodology can also deal with half-hour data. Finally, to validate even more our methodology against the heterogeneity of the groups, we have also created another Global group with 200 smart meters randomly selected from the overall sample. 

As a summary, the first row in Table \ref{summarySM} contains the names of the 18 ACORN groups together with the Global Group, the second and third columns indicate the number of smart meters used by our methodology to train the LSTM model, and to test the forecasting performance, respectively. It should be noticed that no information about the meters in the test set have been used when training the model. We have considered 80\% of the available meters in each group for the training set and the other 20\% of the meters for the testing one. Moreover, the fourth column in Table \ref{summarySM} shows the number of periods (hours) used to train the LSTM whereas the fifth column indicates the number of days the trained LSTM has been used to perform the 24-hours ahead forecasts in our validation (out-of-sample test) scheme. The proposed methodology jointly treats and combines information from all the time series in a given group, hence the estimation window for the LSTM model depends not only on the number of past hours but also on the number of time series. In particular, the larger the size of the group, the smaller the number of hours needed to train the LSTM model. We have considered that the number of hours in the training set is $450000$ hours over the number of meters, with a minimum of 720 hours (roughly one month) and a maximum of 7200 hours (roughly 10 months).

\begin{table}[ht]
	\caption{Summary of the complete dataset}\label{summarySM}
	\centering
	\small
	\begin{tabular}{lrrrr}
		\hline
		Groups & Meters in train  & Meters in  test & Hours in  train & Days in  test\\ 
		\hline
		ACORN-A &  74 &  19 & 5040 & 155 \\ 
		ACORN-B &  14 &   4 & 7200 &  65 \\ 
		ACORN-C &  78 &  19 & 4320 & 185 \\ 
		ACORN-D & 139 &  35 & 2880 & 245 \\ 
		ACORN-E & 801 & 200 & 720 & 335 \\ 
		ACORN-F & 358 &  90 & 720 & 335 \\ 
		ACORN-G & 112 &  28 & 2880 & 245 \\ 
		ACORN-H & 266 &  67 & 1440 & 305 \\ 
		ACORN-I &  30 &   8 & 7200 &  65 \\ 
		ACORN-J &  45 &  11 & 7200 &  65 \\ 
		ACORN-K &  98 &  24 & 3600 & 215 \\ 
		ACORN-L & 181 &  45 & 2160 & 275 \\ 
		ACORN-M &  64 &  16 & 5760 & 125 \\ 
		ACORN-N &  91 &  23 & 3600 & 215 \\ 
		ACORN-O &  55 &  14 & 6480 &  95 \\ 
		ACORN-P &  54 &  14 & 6480 &  95 \\ 
		ACORN-Q & 469 & 117 & 720 & 335 \\ 
		ACORN-U &  22 &   6 & 7200 &  65 \\ 
		Global Group & 160 &  40 & 2160 & 275 \\ 
		\hline
	\end{tabular}
\end{table}
\normalsize

We have also removed smart meters with a high number of missing values (more than 20) and almost zero consumption (standard deviation less than 0.01 kWh). We have replaced the remaining of the missing values by the most recent real observation for each smart meter.
Finally, we have also considered hourly weather data for London area in the same dates as in the smart-meters registers, as can be found in \cite{kaggle}.

In total, our final dataset contains hourly consumption data for 3891 smart meters (for all the 19 groups) along 2013, with a total of 8760 observations for each smart meter. In addition, our dataset also contains temperature and humidity data for each of the 8760 periods, together with extra features (calendar) as the day of the week and the hour in a day, counting a total of 33 features.

\begin{figure}
	\includegraphics[width=6in]{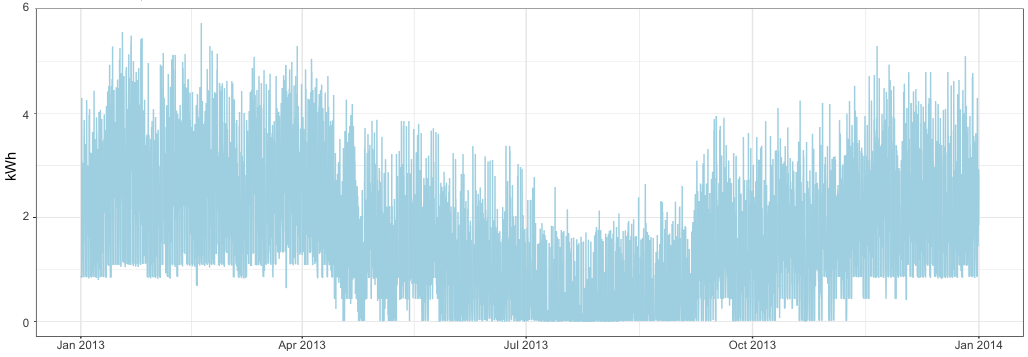}\caption{Consumption for smart-meter 18 along 2013}\label{fig:SM18}
\end{figure}

\begin{figure}
	\includegraphics[width=6in]{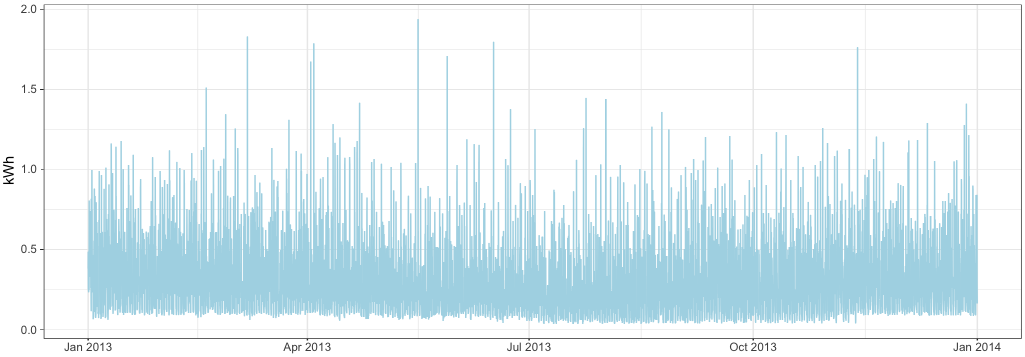}\caption{Consumption for smart-meter 92 along 2013}\label{fig:SM92}
\end{figure}

Next, a summary to understand the behavior of this dataset is provided. To have an idea, we have selected two smart-meters by chance from the ACORN-A group that contains the most affluent people in the UK \emph{lavish lifestyles}. This group contains 93 households but still this group is heterogeneous. In particular, we have selected by chance the 18th and the 92nd households, and in Figures \ref{fig:SM18} and \ref{fig:SM92} we can observe the evolution of their consumption over 2013, respectively. Note the different behavior of these two time-series. Consumer 18th has a higher consumption on average with a higher volatility, but also longer periods with no consumption. On the other hand, 92nd consumer has a lower and more stable consumption, lower but strictly positive. 

\begin{center}
\begin{figure}
	\center \includegraphics[width=5in]{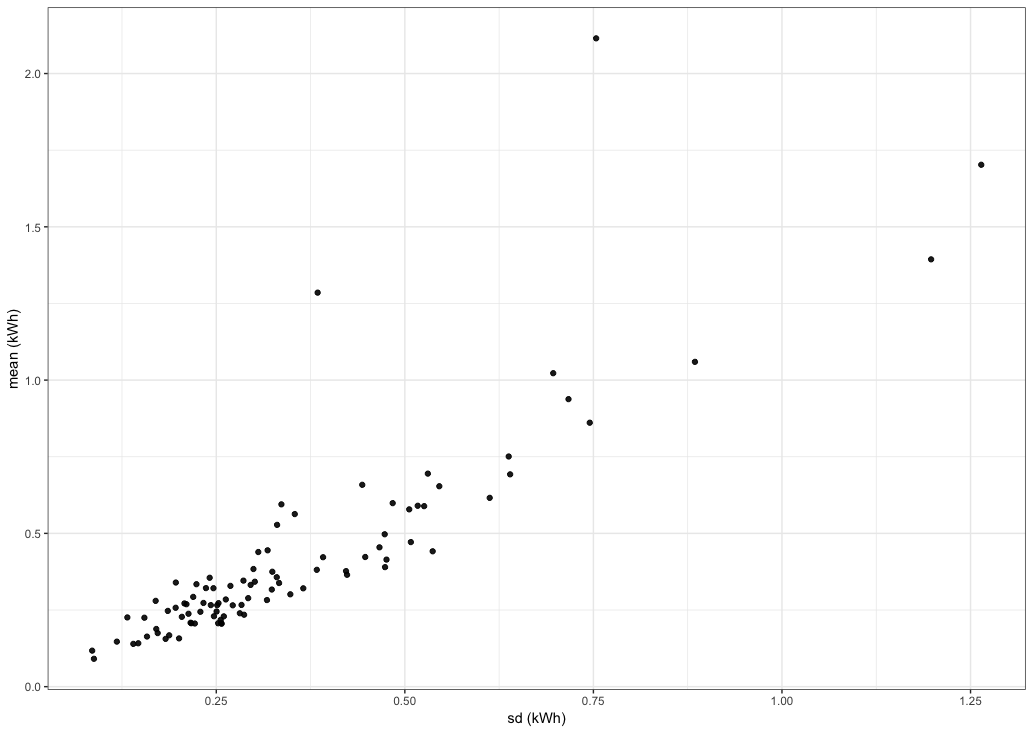}\caption{Average consumption vs standard deviation for each smart meter}\label{fig:MeanSD}
\end{figure}
\end{center}

In Figure \ref{fig:MeanSD}, we can see a scatter plot containing the average consumption in 2013 for each smart meter in ACORN-A versus the corresponding standard deviation. Note the clear linear relationship between consumption and variability, indicating the necessity to normalize the 93 consumptions to make the group more homogeneous.  

Now we will analyze the seasonality. Because we are dealing with hourly data, we expect to have several seasonalities, in particular one of order 24 (daily), other one of order 168 (weekly), and possible one of order 8760 (yearly), but the last one is not observable because we have only data for one year.

\begin{figure}
	\center \includegraphics[width=6in]{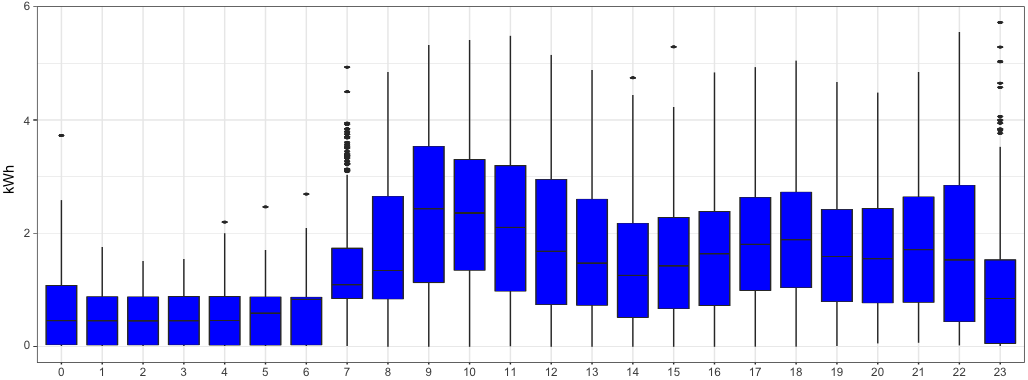}\caption{Daily consumption per hour for smart-meter 18 along 2013}\label{fig:Season18}
\end{figure}

We can analyze the daily seasonality of the different smart meters. For instance, Figure \ref{fig:Season18} reveals a household that decreases consumption during night hours. A similar analysis on weekly seasonality shows that it may also play a relevant role in the time series. Hence, we will consider both the daily and weekly patterns in our methodology. Moreover, from previous figures we can also observe the asymmetric distribution of electricity consumption. This motivates our choice to apply a logarithm transformation to make it more symmetric.

\begin{figure}
	\center \includegraphics[width=6in]{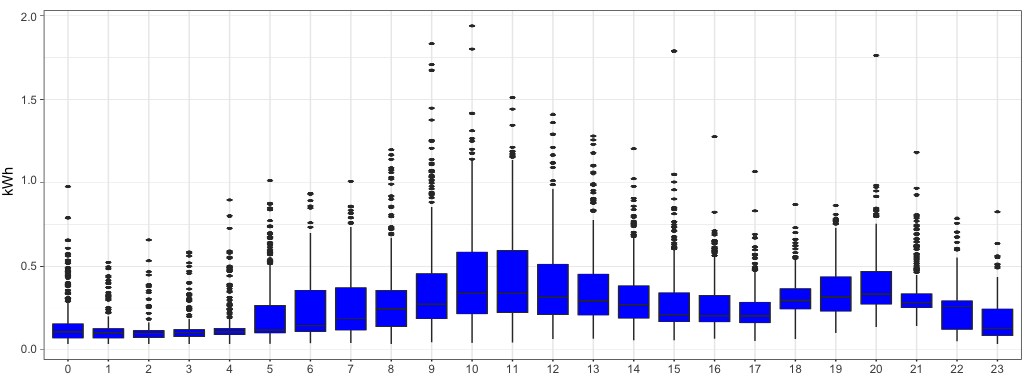}\caption{Daily consumption per hour for smart-meter 92 along 2013}\label{fig:Season92}
\end{figure}

%
%

\begin{figure}
	\center \includegraphics[width=6in]{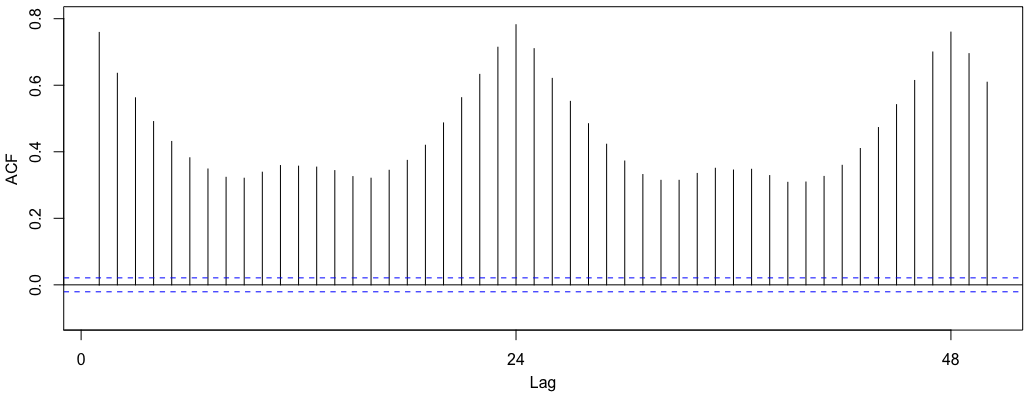}\caption{Autocorrelations for smart-meter 18 along 2013}\label{fig:acf18}
\end{figure}

\begin{figure}
	\center \includegraphics[width=6in]{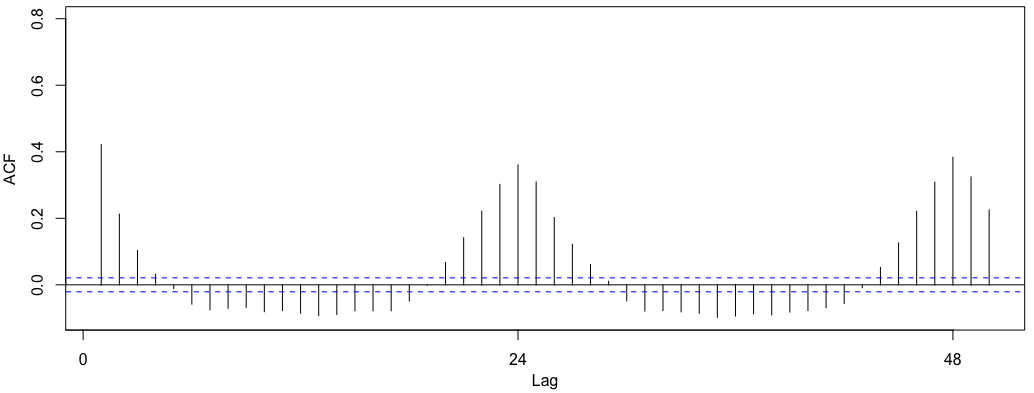}\caption{Autocorrelations for smart-meter 92 along 2013}\label{fig:acf92}
\end{figure}

If we analyze the autocorrelations of household consumptions, useful and different patterns appear. From Figures \ref{fig:acf18} and \ref{fig:acf92} we can observe different patterns of consumption for 18th and 92nd households, respectively. In particular, household 18th presents a high consumption dependency respect to the previous period that decays slowly (long memory). On the other hand, 92nd household presents a small dependency with a higher rate of decay (short memory). In both cases we can observe the dependency respect to previous 24 hours which is an indicator of the daily seasonality previously mentioned.

\begin{figure}
\center \includegraphics[width=6in]{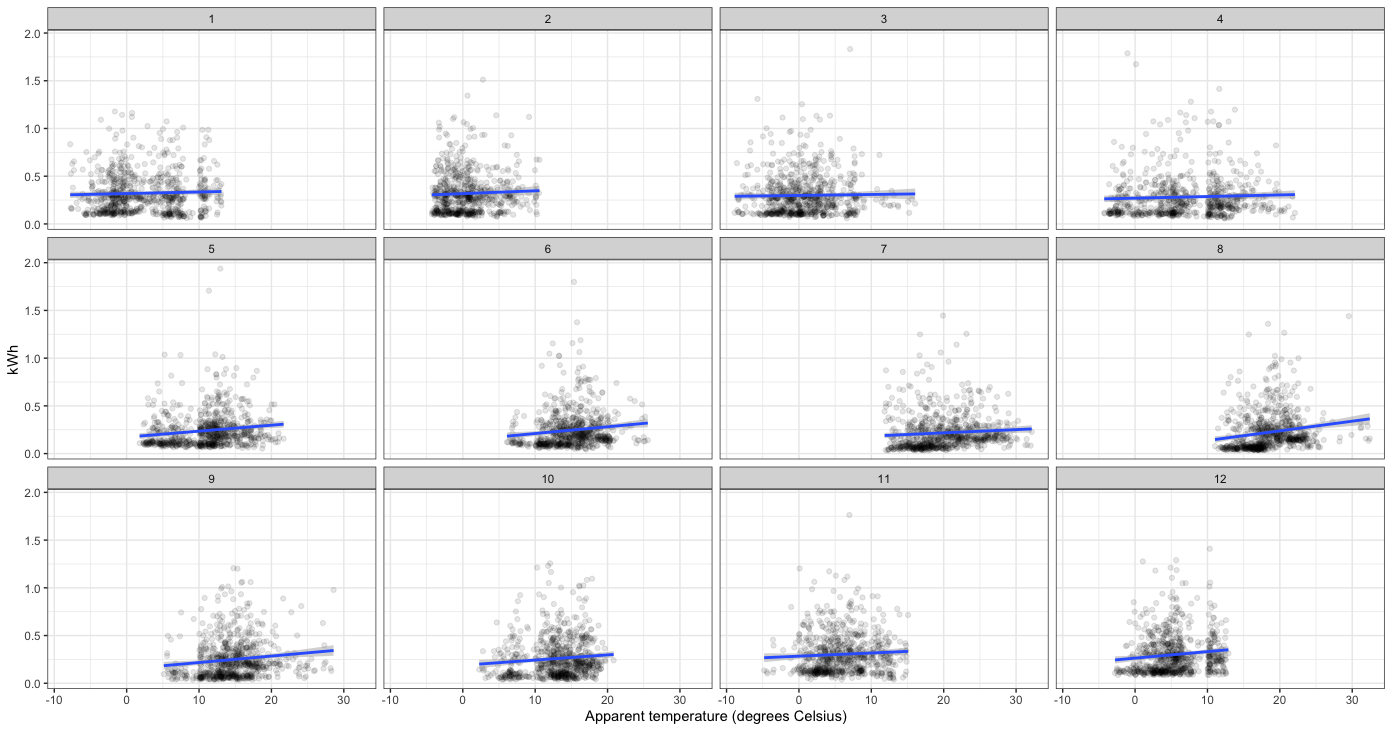}\caption{Consumption for smart-meter 18 vs temperature per month of 2013}\label{fig:Temp18}
\end{figure}

\begin{figure}
\center \includegraphics[width=6in]{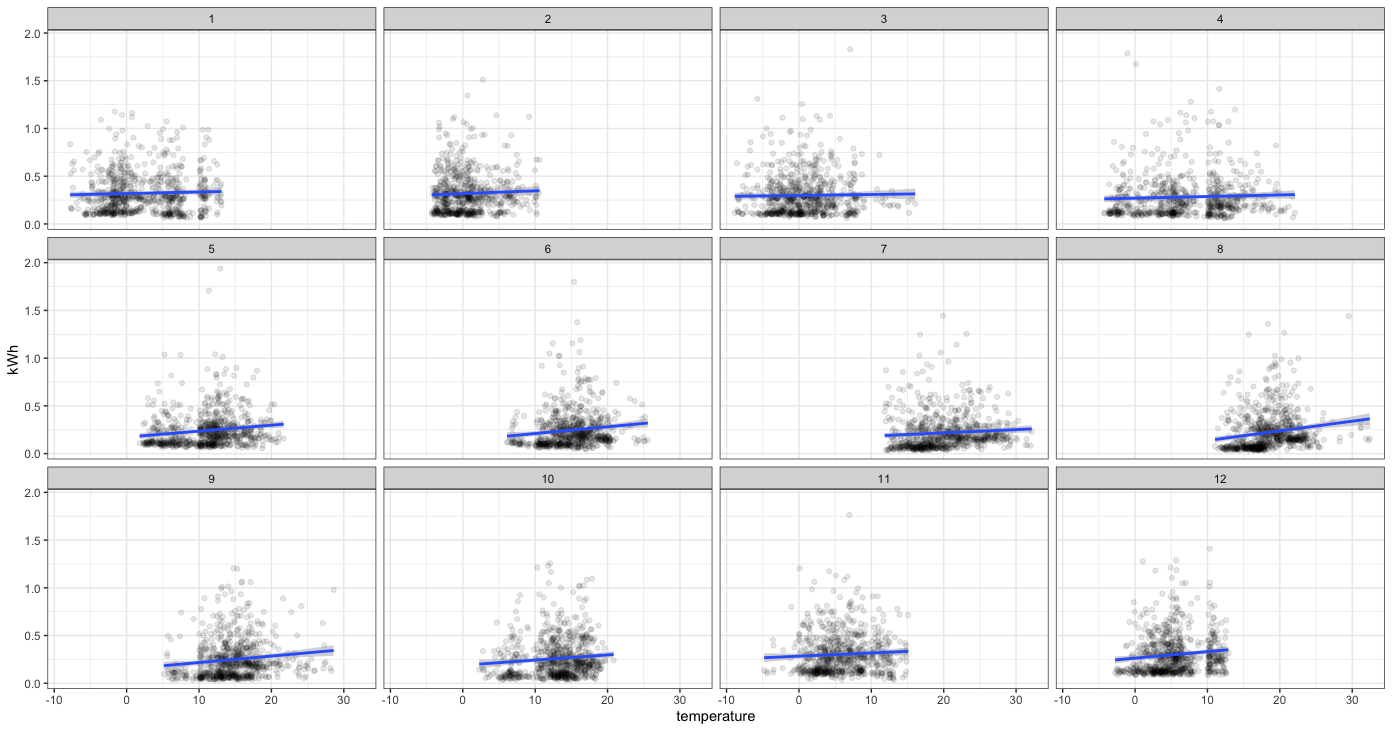}\caption{Consumption for smart-meter 92 vs temperature for each month of 2013}\label{fig:Temp92}
\end{figure}

Next, we analyze the relations between the consumption and meteorological variables. For instance, Figures \ref{fig:Temp18} and \ref{fig:Temp92} show the relation between consumption and temperature for the 18th and 92nd households. We can observe that this relation changes throughout the year as expected, so that it should be considered in the forecasting model.
We have also repeated this analysis to other smart-meters selected by chance from the other 18 groups obtaining similar insights.

As a summary of the descriptive analysis performed in this section, we have observed the collection of time-series analyzed present some complex properties like high-frequency (hourly in our case), non-linearities (especially in relations with meteorological variables), and high volatility (mainly due to the high disaggregation of the consumption). 
This analysis provides the inspiration and motivation for the methodology proposed in the next Section.

\section{Methodology}\label{Met}

The most difficult property to deal with this type of massive time-series is the high volatility, mainly due to the underlying human behavior, that makes the relation between signal and noise unclear. One may use univariate approaches for each series, but this approach is unable to be implemented in practice (maybe millions of models to be trained/estimated) and also does not deal adequately with the non-stationarity in the data.

For this reason, we propose a methodology to deal with a large number of time series. In particular, instead of analyzing each series in isolation, we propose to jointly treat and combine information from several time series contained in a homogeneous group. These group series may be obtained by natural grouping (in our case, the ACORN groups) or statistical clustering (for instance based on features). A review of several clustering techniques to group similar electricity consumers is presented in \cite{chicco2012}.
 
Once the time series are classified into groups, a single RNN with LSTM model is trained over each group. This type of model is specifically designed to capture sequential dependencies between the data, as is the case of time series, and also learn from different consumers dynamics. But note the previous classification is not essential. Out numerical results show for the Global Group, including time series by chance, that accurate results can also be obtained without the previous clustering.  But if a previous classification of the time series into homogeneous groups is available, or a clustering of time series is performed previously to attain homogeneous groups, our methodology is able to exploit better this classification to forecast better each group.


In this Section, we will explain in detail the proposed methodology. As mentioned before, our methodology will be based on ANN and DL because they are flexible and powerful in general, an in particular they can deal with multi-dimensional time series with complex interactions in a natural way. Specifically, we will use RNN for sequential data (time series) which has been proven to work successfully in natural language processing and speech recognition. Recurrent neural networks are a type of neural network where outputs from previous time steps are taken as inputs for the current time step. This creates a network graph or circuit diagram with cycles, which can be unfolded over the input sequence to make predictions over many time steps. 

To train an RNN, the stochastic gradient descent is used, but gradients tend to vanish to zero too early, loosing then the available information. For this reason, LSTM networks were proposed as a type of RNN where the gradients tend to zero slower, improving then the performance. 

\subsection{Framework and Notation}

The main and original idea of our methodology is to train one single model for all the considered time series in a given group. This is another reason to use ANN, because these models tend to perform better as the size of the input increases. Hence, we will join the information from all the time series in a group to train a LSTM model per group. Once the model has been trained using a long history, it can be used to forecast future loads not only for the smart meters considered in the training set but also for new smart meters (see Table \ref{summarySM}).

Next, the main notation to understand our implementation is explained. We will use the TensorFlow  framework \cite{Tensorflow} which is an open source software library for numerical computations written in C++ (and not confined exclusively to ANN). This framework was originally developed by researchers in Google to perform distributed computation, deal with large datasets, automatic differentiation, optimization algorithms, etc.

The main ingredient of the TensorFlow framework is a tensor (multidimensional matrix). As we are dealing with time series, we need a 3-dimensional framework to treat them. The first dimension is the number of samples. In our case, as we are jointly combining the information of many smart meters, the number of samples will be the number of smart meters considered in the training set times the number of periods (hours) in the training set (see second and fourth columns in Table \ref{summarySM}).
The second dimension is the \emph{timesteps}. These are separate time periods of a given variable and for a given observation with influence in future periods. As it was analyzed in the previous section, it was a clear seasonality of order 24 from the daily behavior. For this reason, we have selected timesteps as 24.
Finally, for the third and last dimension we need to consider the features (predictors or regressors) we have available in our dataset. In our case, we have meteorological variables (like apparent temperature and humidity), the aggregated consumption from all the smart meters in each group, 23 dummy variables for the hour of the day, and 6 dummies for the day of the week, summing up to 33 variables for the features dimension (considering also the consumption for each smart meter).

To make the implementation easier, we use Keras \cite{Keras}, which is a high-level neural networks API running on top of TensorFlow. 
The Keras and TensorFlow frameworks allow us to implement and train the LSTM models in an easy way. In particular, we can design the network topology and estimate the weights by minimizing a differentiable loss function through the (mini-batch) gradient descent method, and compute the derivatives using back-propagation (chain’s rule). In particular, we have trained 19 LSTM models, one for each ACORN groups and the additional Global Group, with around 12,000 weights each of them that need to be estimated along a highly non-linear function from the loss function, that compares the forecasts with the real values.

\subsection{Data Preparation}

Before applying the recurrent neural network model, it is needed to perform some data pre-processing. This step usually needs to deal with missing values and outliers, make some feature extraction, scale and normalize the data, etc.
In our case, as mentioned in Section \ref{DA}, we have replaced the missing values by the most recent observation in each smart meter. Then, we proceed to make the consumption more symmetric and take the natural logarithm as the first transformation. Later, we normalize the consumption in order to have the same mean and variance for all the households. These transformations are useful to make the time series more stationary, implying better performance for the LSTM model.
However, we do not transform the data to remove the seasonality because the LSTM model is able to capture it in a natural way through the \emph{timesteps} defined previously.

Finally, to use TensorFlow we need to convert the original input data into 3-dimensional tensors for time-series. To do that, we need to build a design matrix and a target matrix. The design matrix is a tensor containing the information from the past that the LSTM will use to forecast the future. In our case, it will be a 3-dimensional array where the first dimension is $N$*$T$, being $N$ the number of smart meters and $T$ the number of periods considered in the past, i.e. the second and fourth columns in Table \ref{summarySM}. The second dimension is the timesteps, 24 in our case, denoting the number of recent observations the LSTM will be used to forecast next hours. Finally, the last dimension is the number of features, 33 in our case as explained before. 
The target matrix is again a 3-dimensional array where the first dimension is $N$ because we are interested in forecasting next period, the second dimension is the forecasting horizon for the next period, 24 in our case, and the last dimension is again 33. 
An example of one row of the design matrix for one smart meter would be:
\[
  z_1, z_2, \ldots, z_{24}
\]
where each value denotes past consumption for a given hour. An example of the same row for the target matrix and the same smart meter would be:
\[
z_{25}, z_{26}, \ldots, z_{48}
\]

Note we need to build these arrays for each of the available features.
Next, the details of the LSTM model will be explained.

\subsection{The Proposed LSTM Model}

The LSTM model in Keras is defined as a sequence of layers. In our experiments, all the LSTM models are trained with the same network topology as follows. The first layer in the network defines the 3-dimensional units of the tensors as explained before. This layer is a LSTM one containing 32 units or number of neurons. Then we need to define the activation function to transform the output from each unit as the input for the next layer. The choice of this activation function is important and will affect the forecasting performance. We have used the \textit{hyperbolic tangent} as the activation function, but others can be used as the \textit{sigmoid} or \textit{softmax} ones.
Then, multiple layers can be stacked by adding them to the sequential process. In our case,  we have added another LSTM layer with 16 units, and finally a dense layer (or fully connected one) with 24 units that will provide the corresponding 24-hours ahead forecast.

Moreover, to avoid overfitting in such a large network and improve performance, we need to add some type of regularization or dropout. If we choose a regularization approach, it reduces overfitting by adding some bias to the estimation while reducing the variance. On the other hand, if we choose dropout, it reduces overfitting by randomly making zero some units in the layer during the training steps. We have chosen this last approach. In particular, in the first recurrent layer we have selected a dropout rate of 10\% of the recurrent units followed by a dropout rate of 10\% for the input units of the layer. For the last LSTM layer, we have selected a dropout rate of 5\%, both for the recurrent units and for the input ones.

Once we have designed the topology of the network, we need to compile it to make the estimation more efficient. To do that, we need to select the loss function and the optimization algorithm to train the network. For the loss function we have selected the \textit{mean absolute value}, and for the optimization algorithm we have selected the \textit{Adam} optimizer, which is a version of the stochastic gradient method, with a learning rate of $0.001$.

All the previous hyper-parameters have been selected after trying other network architecture and values and observing the out-of-sample loss function.
After the network is compiled for each the 19 LSTM models, we can proceed to fit or optimize the associated weights. This is the most expensive step in the methodology from a computational point of view. For this reason, we propose to train the network only once in a year, with a large number of time series and observations, and use the trained LSTM to perform the 24-hours ahead forecasts for all the desired periods and smart meters in the future. This last forecasting step can be executed very fast.

The fitting or optimization process to train the network uses the back-propagation algorithm, together with the optimization algorithm and loss function defined previously. The back-propagation algorithm requires to define the number of epochs or times the optimization algorithm uses the complete training set. In our case, we have selected 40 epochs.
Each epoch can be partitioned into a fixed-sized number of rows from the training set, called batch. This subset of the training set will improve the performance of the optimization algorithm. We have selected a batch size of 1000 hours (around 40 days).
Once the 19 LSTM models have been trained, they are ready to perform the forecasts for all groups as the next section shows.

\subsection{The back-testing}

In this subsection, we explain how we have developed the back-testing scheme to evaluate the performance of the proposed methodology. First, we have information about 3891 household consumptions for all hours in 2013, i.e. 8760 hours. We have organized these 3891 smart meters in 18 groups according to the ACORN classification, as summarized in Table \ref{summarySM}, plus the Global Group with 200 smart meters selected by chance out of the previous 18 groups. For each group, we train a LSTM (as explained in previous section) using the number of periods indicated in the fourth column in Table \ref{summarySM}, leaving the last days in the sample (fifth column in Table \ref{summarySM}) to test the 24-hours ahead forecasts for each of these days. Moreover, for each group, we have selected around 80\% of the smart meters to train the models (as indicated in the second column of Table \ref{summarySM}), leaving the other 20\% (third column in Table \ref{summarySM}) to test the forecasts.
That means, for the out-of-sample evaluation of the models, we consider two dimensions: a time direction with out-of-sample periods, and a smart-meter direction for out-of-sample households (as they were new customers).

Besides the proposed LSTM model, we have used two well-known benchmarks: i) a seasonal \textit{naive} approach, and ii) an \textit{auto.arima} approach, as described in \cite{hyndman2008}.
We have also tried a \textit{tbats} method, as described in \cite{de2011forecasting}, but this method becomes very unstable for some of the time series providing bad results in practice.
For the seasonal \textit{naive} approach, we forecast the next 24-hours ahead consumptions, with the last 24-hour ones. This is one of the most successful approaches in practice, because it can be implemented and executed with little computational effort, it is stable and scalable, and provides reasonable performance for massive time-series.

For the \textit{auto.arima} approach, note this is an automatic framework for univariate time-series. Its performance is reasonable in general, but it does not deal well with possible non-stationarity in the data and non-linearities. Moreover, this approach is unable to implement in practice for a big energy utility because it requires the training of maybe millions of models, one for each available customer. In any case, we have implemented this approach for all the 3891 smart meters in order to analyze and compare better the results.
 
Finally, we have obtained results from our proposed LSTM model. This model requires in practice training only once or twice a year and then, individual forecasts can be obtained with little computational costs. Because we train only one model for all the time series in a group at once, our proposal has a great potential for large scale applications. 

In the next subsection, the numerical results obtained by implementing previous approaches are shown and discussed. To do that, we have computed the mean-absolute error (MAE) for each approach, and for each out-of-sample period and smart meter. In particular, if $z_{n,t+h}$ denotes the real consumption of smart meter $n$, for period $t+h$, and for horizon $h=1,\ldots,24$, then the MAE for the $i$-th approach, $\hat{z}_{i,n,t+h}$ is defined as:
\[
  \text{MAE}(i,n,t) = \frac{1}{24}\sum_{h=1}^{24} |\hat{z}_{i,n,t+h}-z_{n,t+h}|,
\]
for each of the out-of-sample periods $t=1, \ldots, T'$, and each of the smart meters $n=1,\ldots,N'$.

\begin{figure}\centering 
	\includegraphics[width=6in]{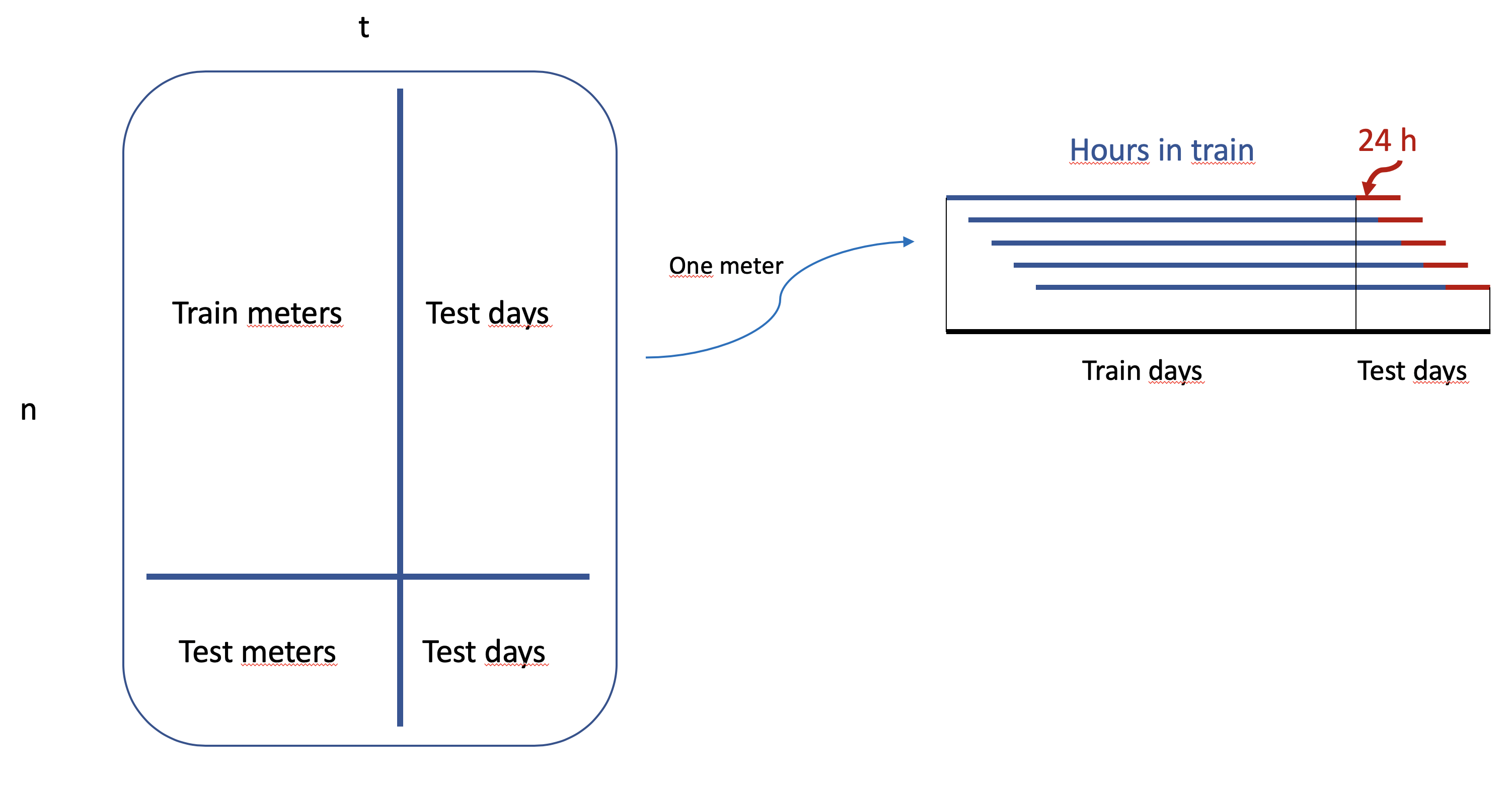}\caption{Scheme of out-of-sample evaluation along meters and time for each group and method}\label{fig:scheme}
\end{figure}

Finally, in next Subsection median MAEs are computed for each group and method: the median MAE in testing meters corresponds to the median of MAE along the out-of-sample periods $t$ and then the median along the testing meters $n$, whereas the median MAE in testing days corresponds to the median of MAE along the meters $n$ and then the median along the out-of-sample periods $t$. Figure \ref{fig:scheme} illustrates the performance scheme.

\subsection{Results}

\begin{figure}
	\center \includegraphics[width=6in]{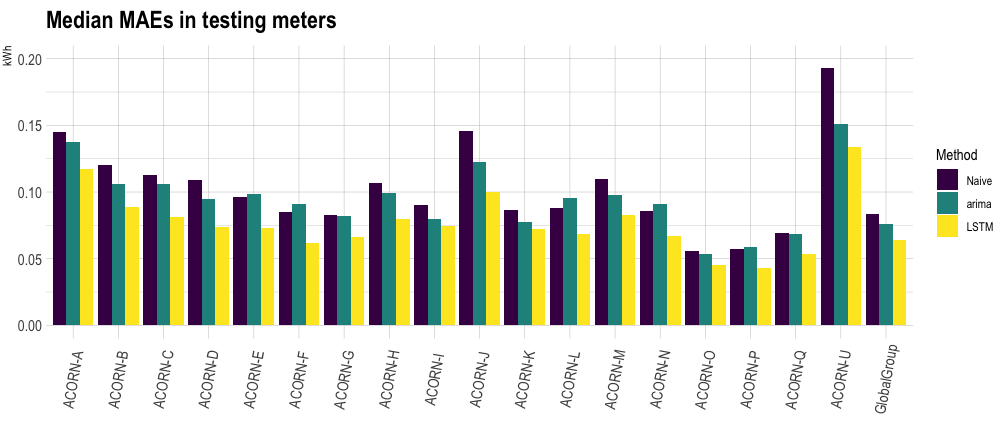}\caption{Out-of-sample performance for smart meters in the testing set}\label{fig:TestMeters}
\end{figure}

In this section, we summarize the main numerical results obtained when applying the three approaches (\textit{naive}, \textit{auto.arima}, and the proposed LSTM model) for each of the 19 groups considered. In Figure \ref{fig:TestMeters}, we can observe the median MAEs for the three implemented approaches along the testing meters. In particular, for each of the smart meters in the testing set for each group, we have computed the median error for all the MAEs obtained in the out-of-sample days in the back-testing. It can be noted how the \textit{auto.arima} approach is on average a 5\% better than the \textit{naive} approach but sometimes is a bit worse. On the other hand, the LSTM approach attains the best performance. In particular, the performance of the LSTM approach is around 19\% better than that of the \textit{auto.arima} and around 24\% better than that of the \textit{naive} approach. That implies a good performance of the proposed methodology even for new customers in the database.

\begin{figure}
	\center \includegraphics[width=6in]{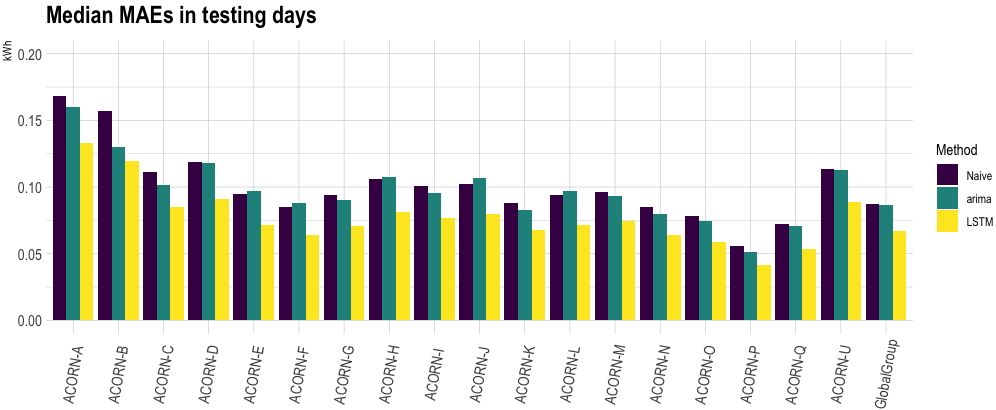}\caption{Out-of-sample performance for testing days}\label{fig:TestPeriods}
\end{figure}

On the other hand, Figure \ref{fig:TestPeriods} shows the median MAEs for all smart meters along the out-of-sample days. Note the errors of \textit{auto.arima} and \textit{naive} approaches behave similarly along time (\textit{auto.arima} performs around 3\% than the \textit{naive}), while the proposed LSTM outperforms the \textit{naive} approach by 24\% on average and the \textit{auto.arima} one by 21\% on average over \emph{all} the out-of-sample periods.

\begin{figure}
	\center \includegraphics[width=6in]{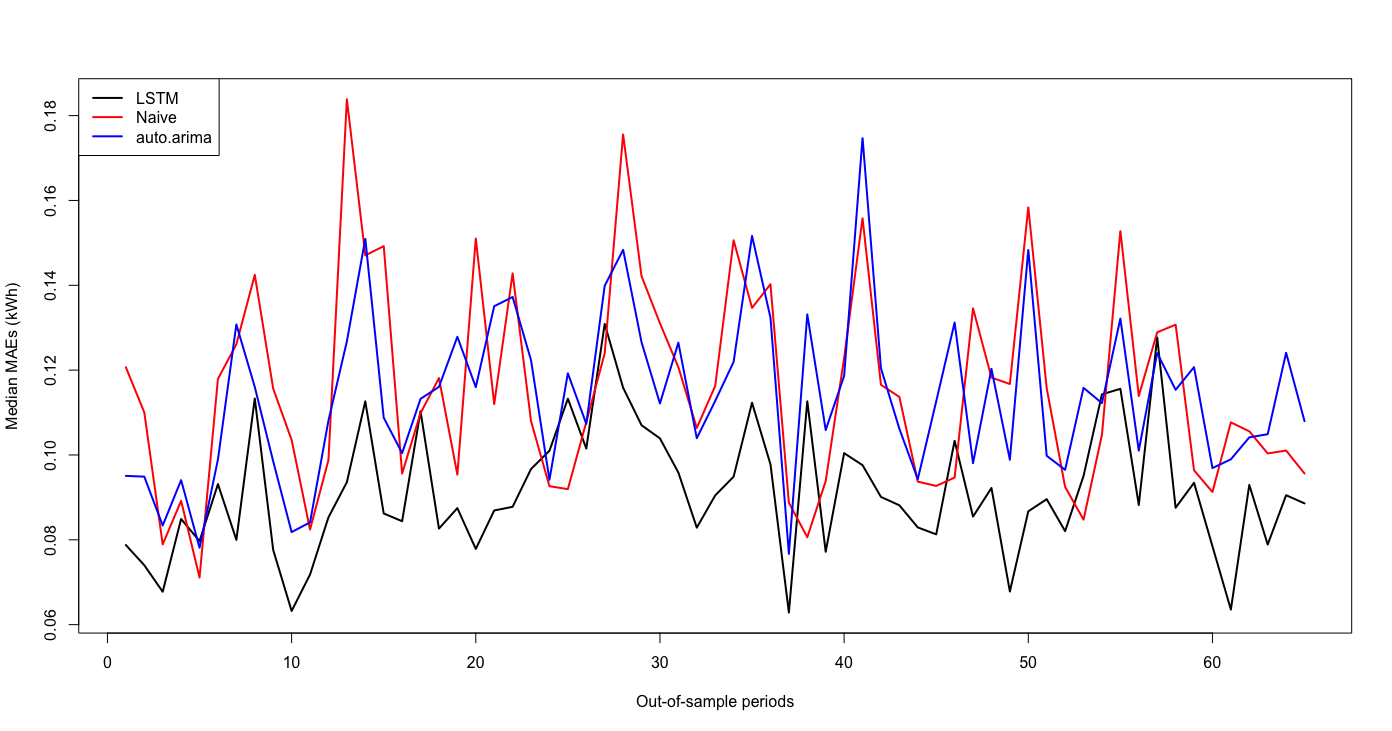}\caption{Out-of-sample performance for testing days (ACORN-U group)}\label{fig:ACORNUpath}
\end{figure}

To have a better idea about the forecasted load profiles by the three considered methods, we have selected the ACORN-U group because it has the largest error in Figure \ref{fig:TestMeters}. In particular,  Figure \ref{fig:ACORNUpath} shows the evolution of the median MAEs along the 65 out-of-sample days for all the meters in the ACORN-U group. Note that although forecasting performance depends on time, the proposed LSTM procedure has consistently better out-of-sample errors than the benchmarks.

\begin{figure}
	\center \includegraphics[width=6in]{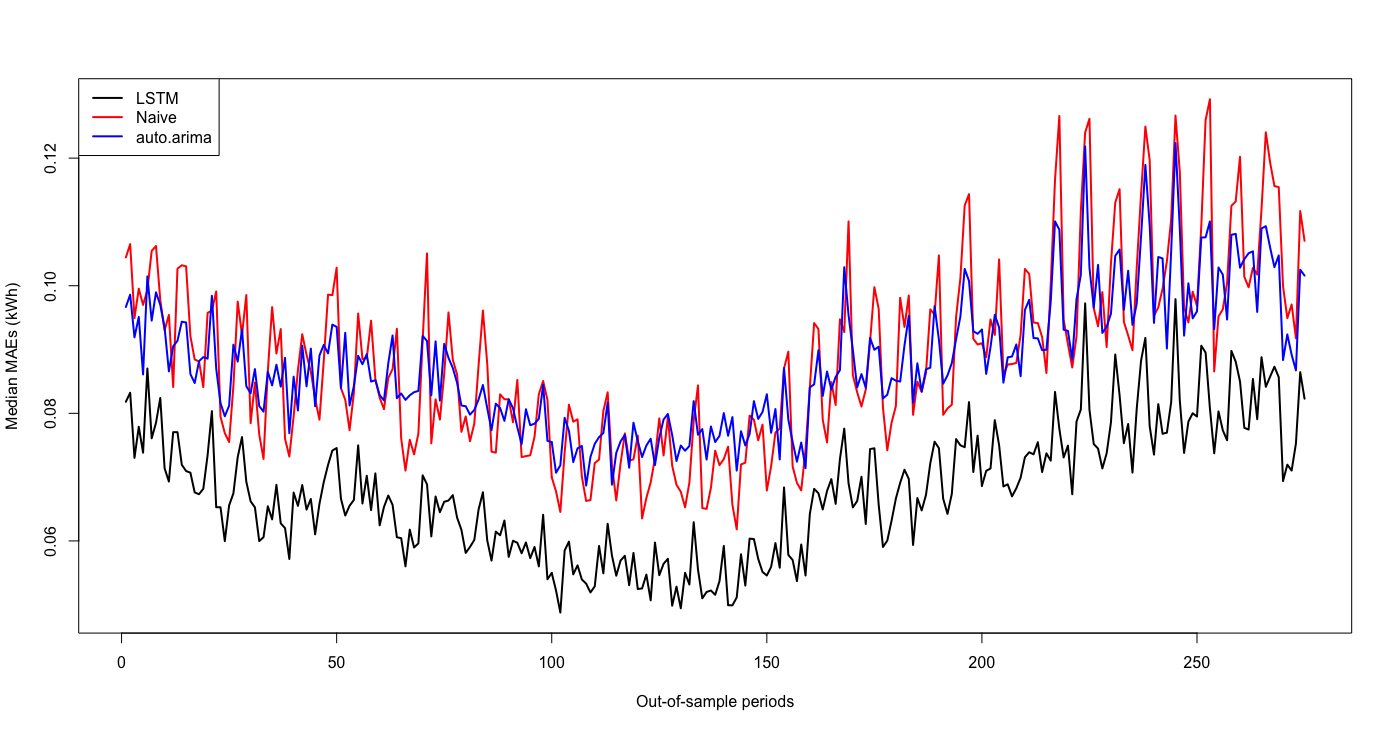}\caption{Out-of-sample performance for testing days (Global Group)}\label{fig:GlobalGrouppath}
\end{figure}

Finally, It is worth analyzing the behavior of our our proposal for the Global Group. This group contains 200 time series selected by chance from the other 18 groups. We have chosen 160 of those series to train the unique LSTM and used it to forecast the other 40 series in the testing set. Figure \ref{fig:GlobalGrouppath} shows the evolution of the median MAEs along the 275 days in the out-of-sample period for those 40 meters. Again, our method outperforms consistently the two benchmarks by more than 20\%, implying the effectiveness of the proposal even when a classification of groups is not available a priori. Of course, if a previous classification of the time series into homogeneous groups is available, our methodology is able to exploit better this classification to forecast better each group. 

The evaluation of these models indicates our proposal have achieved promising and competitive results for all the geo-demographic groups considered in our dataset, implying a good potential for its implementation in large-scale smart-meter load forecasting.

\section{Conclusions}\label{Con}

We have proposed a general methodology, based on recurrent neural networks and specifically a LSTM model, that is able to forecast consumptions from electricity smart-meters in an efficient way for a utility. These forecasts could be used in the implementation of demand response policies, to anticipate and prevent systems peaks or congested lines, to adapt load consumption to renewable generation availability, to identify consumption anomalies that may originate from equipment failure or from electricity theft, etc. 

Instead of using traditional and univariate approaches for each smart meter, we propose a single but complex LSTM model that captures the main features of individual consumptions and also information from different household consumptions. As a consequence, the model attains promising results respect to competitive benchmarks, more than 20\% better performance on average along all the testing periods and all the testing meters in our back-testing experiment.

We would like to mention some disadvantages of the proposed approach. As any other complex neural-network, there are some difficulties in the design of the network. In particular, there is no a clear way to design the network topology about the configuration of the layers and the corresponding units or neurons. And there are many hyper-parameters to be considered, like the number of layers, the number of units, the activation functions, the type of regularization or dropout, optimization parameters, etc. That implies a great effort and time before the network starts working properly. In addition, neural networks in general are difficult to understand and interpret.

Moreover, the competitive good results depend on the degree of homogeneity of the considered smart meters. In this work, we have used 19 groups of households, the smallest one with 18 smart meters and the biggest one 1001 ones. We have obtained promising results for the near 4000 smart meters, even when the groups have some degree of heterogeneity. But we do not expect good results for smart meters with very unusual profiles (outliers) in its group. 
On the other hand, the success of the proposed model is partially explained by the large amount of data used to train the model. Because we are jointly treating and combining the information of all the available time series in a group, the network is able to capture non-linear relations, seasonalities and other hidden patterns that other traditional approaches cannot capture because lack of enough data.

As a summary, our methodology is able to outperform competitive univariate forecasting tools for electricity consumption, providing an implementable and scalable approach for massive time-series forecasting. In particular, it may provide near real-time forecast for hundreds of thousands of smart meters. The model training would need to be done only once every few months, and by using just a representative subset of the smart meters time series. Once the network is trained, and its input properly defined, it can provide short-term smart meters’ forecast at almost no computational costs.

%

\section*{Acknowledgment}

The authors gratefully acknowledge the financial support from the Spanish government through projects
MTM2017-88979-P and ECO2015-66593-P, and from Fundación Iberdrola through ``Ayudas a la Investigación en Energía y Medio Ambiente 2018''.

\bibliographystyle{IEEEtran}
\bibliography{forecasting_LSTM_arxiv_II}

\begin{thebibliography}{10}
\providecommand{\url}[1]{#1}
\csname url@samestyle\endcsname
\providecommand{\newblock}{\relax}
\providecommand{\bibinfo}[2]{#2}
\providecommand{\BIBentrySTDinterwordspacing}{\spaceskip=0pt\relax}
\providecommand{\BIBentryALTinterwordstretchfactor}{4}
\providecommand{\BIBentryALTinterwordspacing}{\spaceskip=\fontdimen2\font plus
\BIBentryALTinterwordstretchfactor\fontdimen3\font minus
  \fontdimen4\font\relax}
\providecommand{\BIBforeignlanguage}[2]{{%
\expandafter\ifx\csname l@#1\endcsname\relax
\typeout{** WARNING: IEEEtran.bst: No hyphenation pattern has been}%
\typeout{** loaded for the language `#1'. Using the pattern for}%
\typeout{** the default language instead.}%
\else
\language=\csname l@#1\endcsname
\fi
#2}}
\providecommand{\BIBdecl}{\relax}
\BIBdecl

\bibitem{depuru2011smart}
S.~S. S.~R. Depuru, L.~Wang, V.~Devabhaktuni, and N.~Gudi, ``Smart meters for
  power grid—challenges, issues, advantages and status,'' in \emph{2011
  IEEE/PES Power Systems Conference and Exposition}.\hskip 1em plus 0.5em minus
  0.4em\relax IEEE, 2011, pp. 1--7.

\bibitem{yildiz2017recent}
B.~Yildiz, J.~Bilbao, J.~Dore, and A.~Sproul, ``Recent advances in the analysis
  of residential electricity consumption and applications of smart meter
  data,'' \emph{Applied Energy}, vol. 208, pp. 402--427, 2017.

\bibitem{wang2018review}
Y.~Wang, Q.~Chen, T.~Hong, and C.~Kang, ``Review of smart meter data analytics:
  {A}pplications, methodologies, and challenges,'' \emph{IEEE Transactions on
  Smart Grid}, vol.~10, no.~3, pp. 3125--3148, 2018.

\bibitem{conejo2010decision}
A.~J. Conejo, M.~Carri{\'o}n, J.~M. Morales \emph{et~al.}, \emph{Decision
  making under uncertainty in electricity markets}.\hskip 1em plus 0.5em minus
  0.4em\relax Springer, 2010, vol.~1.

\bibitem{hong2016probabilistic}
T.~Hong and S.~Fan, ``Probabilistic electric load forecasting: A tutorial
  review,'' \emph{International Journal of Forecasting}, vol.~32, no.~3, pp.
  914--938, 2016.

\bibitem{bandara2017forecasting}
K.~Bandara, C.~Bergmeir, and S.~Smyl, ``Forecasting across time series
  databases using recurrent neural networks on groups of similar series: {A}
  clustering approach,'' \emph{arXiv preprint arXiv:1710.03222}, 2017.

\bibitem{hochreiter1997long}
S.~Hochreiter and J.~Schmidhuber, ``Long short-term memory,'' \emph{Neural
  computation}, vol.~9, no.~8, pp. 1735--1780, 1997.

\bibitem{edwards2012predicting}
R.~E. Edwards, J.~New, and L.~E. Parker, ``Predicting future hourly residential
  electrical consumption: {A} machine learning case study,'' \emph{Energy and
  Buildings}, vol.~49, pp. 591--603, 2012.

\bibitem{raza2015review}
M.~Q. Raza and A.~Khosravi, ``A review on artificial intelligence based load
  demand forecasting techniques for smart grid and buildings,'' \emph{Renewable
  and Sustainable Energy Reviews}, vol.~50, pp. 1352--1372, 2015.

\bibitem{ma2017modeling}
W.~Ma, S.~Fang, G.~Liu, and R.~Zhou, ``Modeling of district load forecasting
  for distributed energy system,'' \emph{Applied Energy}, vol. 204, pp.
  181--205, 2017.

\bibitem{gajowniczek2017electricity}
K.~Gajowniczek and T.~Zabkowski, ``Electricity forecasting on the individual
  household level enhanced based on activity patterns,'' \emph{PloS one},
  vol.~12, no.~4, p. e0174098, 2017.

\bibitem{hsiao2014household}
Y.-H. Hsiao, ``Household electricity demand forecast based on context
  information and user daily schedule analysis from meter data,'' \emph{IEEE
  Transactions on Industrial Informatics}, vol.~11, no.~1, pp. 33--43, 2014.

\bibitem{sevlian2018scaling}
R.~Sevlian and R.~Rajagopal, ``A scaling law for short term load forecasting on
  varying levels of aggregation,'' \emph{International Journal of Electrical
  Power \& Energy Systems}, vol.~98, pp. 350--361, 2018.

\bibitem{taieb2016forecasting}
S.~B. Taieb, R.~Huser, R.~J. Hyndman, and M.~G. Genton, ``Forecasting
  uncertainty in electricity smart meter data by boosting additive quantile
  regression,'' \emph{IEEE Transactions on Smart Grid}, vol.~7, no.~5, pp.
  2448--2455, 2016.

\bibitem{taieb2019hierarchical}
S.~B. Taieb, J.~W. Taylor, and R.~J. Hyndman, ``Hierarchical probabilistic
  forecasting of electricity demand with smart meter data,'' 2019.

\bibitem{li2017sparse}
P.~Li, B.~Zhang, Y.~Weng, and R.~Rajagopal, ``A sparse linear model and
  significance test for individual consumption prediction,'' \emph{IEEE
  Transactions on Power Systems}, vol.~32, no.~6, pp. 4489--4500, 2017.

\bibitem{quilumba2014using}
F.~L. Quilumba, W.-J. Lee, H.~Huang, D.~Y. Wang, and R.~L. Szabados, ``Using
  smart meter data to improve the accuracy of intraday load forecasting
  considering customer behavior similarities,'' \emph{IEEE Transactions on
  Smart Grid}, vol.~6, no.~2, pp. 911--918, 2014.

\bibitem{chitsaz2015short}
H.~Chitsaz, H.~Shaker, H.~Zareipour, D.~Wood, and N.~Amjady, ``Short-term
  electricity load forecasting of buildings in microgrids,'' \emph{Energy and
  Buildings}, vol.~99, pp. 50--60, 2015.

\bibitem{tascikaraoglu2016short}
A.~Tascikaraoglu and B.~M. Sanandaji, ``Short-term residential electric load
  forecasting: A compressive spatio-temporal approach,'' \emph{Energy and
  Buildings}, vol. 111, pp. 380--392, 2016.

\bibitem{yildiz2018household}
B.~Yildiz, J.~I. Bilbao, J.~Dore, and A.~Sproul, ``Household electricity load
  forecasting using historical smart meter data with clustering and
  classification techniques,'' in \emph{2018 IEEE Innovative Smart Grid
  Technologies-Asia (ISGT Asia)}.\hskip 1em plus 0.5em minus 0.4em\relax IEEE,
  2018, pp. 873--879.

\bibitem{ahmad2019deep}
T.~Ahmad and H.~Chen, ``Deep learning for multi-scale smart energy
  forecasting,'' \emph{Energy}, vol. 175, pp. 98--112, 2019.

\bibitem{mocanu2016deep}
E.~Mocanu, P.~H. Nguyen, M.~Gibescu, and W.~L. Kling, ``Deep learning for
  estimating building energy consumption,'' \emph{Sustainable Energy, Grids and
  Networks}, vol.~6, pp. 91--99, 2016.

\bibitem{shi2017deep}
H.~Shi, M.~Xu, and R.~Li, ``Deep learning for household load forecasting—a
  novel pooling deep {RNN},'' \emph{IEEE Transactions on Smart Grid}, vol.~9,
  no.~5, pp. 5271--5280, 2017.

\bibitem{kong2017short}
W.~Kong, Z.~Y. Dong, Y.~Jia, D.~J. Hill, Y.~Xu, and Y.~Zhang, ``Short-term
  residential load forecasting based on {LSTM} recurrent neural network,''
  \emph{IEEE Transactions on Smart Grid}, vol.~10, no.~1, pp. 841--851, 2017.

\bibitem{wang2019probabilistic}
Y.~Wang, D.~Gan, M.~Sun, N.~Zhang, Z.~Lu, and C.~Kang, ``Probabilistic
  individual load forecasting using pinball loss guided {LSTM},'' \emph{Applied
  Energy}, vol. 235, pp. 10--20, 2019.

\bibitem{dataLondon}
\BIBentryALTinterwordspacing
 [Online]. Available:
  \url{https://data.london.gov.uk/dataset/smartmeter-energy-use-data-in-london-households}
\BIBentrySTDinterwordspacing

\bibitem{CACI}
\BIBentryALTinterwordspacing
 [Online]. Available:
  \url{https://acorn.caci.co.uk/downloads/Acorn-User-guide.pdf}
\BIBentrySTDinterwordspacing

\bibitem{kaggle}
\BIBentryALTinterwordspacing
 [Online]. Available:
  \url{https://www.kaggle.com/jeanmidev/smart-meters-in-london}
\BIBentrySTDinterwordspacing

\bibitem{chicco2012}
G.~Chicco, ``Overview and performance assessment of the clustering methods for
  electrical load pattern grouping,'' \emph{Energy}, vol.~42, no.~1, 2012.

\bibitem{Tensorflow}
\BIBentryALTinterwordspacing
 [Online]. Available: \url{https://www.tensorflow.org}
\BIBentrySTDinterwordspacing

\bibitem{Keras}
\BIBentryALTinterwordspacing
 [Online]. Available: \url{https://keras.io}
\BIBentrySTDinterwordspacing

\bibitem{hyndman2008}
R.~Hyndman and Y.~Khandakar, ``Automatic time series forecasting: The forecast
  package for {R},'' \emph{Journal of Statistical Software}, vol.~26, 2008.

\bibitem{de2011forecasting}
A.~M. De~Livera, R.~J. Hyndman, and R.~D. Snyder, ``Forecasting time series
  with complex seasonal patterns using exponential smoothing,'' \emph{Journal
  of the American Statistical Association}, vol. 106, no. 496, pp. 1513--1527,
  2011.

\end{thebibliography}

\end{document}